# Mapping Pipelines and Simultaneous Localization for Petrochemical Industry Robots

1st Mahta Akhyani

University of Tehran

Tehran, Iran Mahtaakhyani@gmail.com

*Abstract—* Inspecting petrochemical pipelines is challenging due to hazardous materials, narrow diameters, and inaccessible locations. Mobile robots are promising for autonomous pipeline inspection and mapping. This project aimed to simulate and implement a robot capable of simultaneous localization and mapping (SLAM) in an indoor maze-like environment representing simplified pipelines. The approach involved simulating a differential drive robot in Gazebo/ROS, equipping it with sensors, implementing SLAM using mapping, and path planning with move_base.

A physical robot was then built and tested by manually driving it in a constructed maze while collecting sensor data and mapping. Sensor fusion of wheel encoders, Kinect camera, and inertial measurement unit (IMU) data was explored to improve odometry and mapping accuracy without encoders. The final map had reasonable correspondence to the true maze despite lacking wheel encoders. In summary, results show the feasibility of using ROS-based SLAM for pipeline inspection if accounting for real-world complexities.

## I. Introduction

Inspecting petrochemical pipelines is an extremely important but challenging task. Petrochemical pipelines form expansive networks across sites to transport various hazardous materials under diverse operating conditions. The transported substances often have highly corrosive and flammable properties. At the same time, the pipelines are subject to internal and external stresses and damage from oxidation and corrosion of the pipe surface. This can sometimes lead to hazardous leakage. Therefore, the pipelines need to undergo frequent inspection on a regular basis.

Unfortunately, pipeline inspection is commonly recognized as a slow, tedious, and labor-intensive process. In most cases, the pipelines are not readily accessible by humans, and sometimes entire refinery operations have to be stopped during inspection. Monitoring is thus difficult. Since many internal sections are not reachable by humans, inspection often has to be done externally, which involves draining insulating substances again. Sometimes, the entire pipeline is replaced even for a minor defect, escalating maintenance costs. Replacing just the damaged section sometimes requires excavating massive stretches of the region. In such situations, knowing the location of the damaged part is critical.

In light of these challenges, using various mobile robots for pipeline inspection systems is gaining increasing importance, as they can conduct faster and more precise inspections. Given the continuous flow of diverse fluids under varying operational conditions, these pipelines require regular checks. These pipelines usually have small diameters, making human inspection extremely difficult. Therefore, building an independent pipe inspection robot is imperative.

In this regard, the aim of this thesis is to study and implement a SLAM algorithm by simplifying the pipelines to a maze-like environment that can be experimentally evaluated, so the robot can independently move in different parts of the maze and perform the inspection process by receiving and mapping the hypothetical pipeline walls or maze boundaries. The first goal is to simulate a simple robot to implement a 2D SLAM algorithm in a custom indoor maze environment in Gazebo, a ROS-based 3D robotics simulator framework. The next goal is evaluating the generated map based on quality, robot localization error, and improving localization accuracy and map quality. Additionally, this process is tested on an actual physical robot, which is a differential drive mobile robot, and is simulated.

Mobile robots equipped with SLAM capabilities have promising applications for autonomous inspection of hazardous industrial environments like petrochemical pipelines. This chapter provides background on the importance and challenges of pipeline inspection, and how robotic systems can help address these challenges.

Petrochemical pipelines transport a range of hazardous substances including flammable, corrosive, and toxic materials. Regular inspection is critical to detect any leaks or damage before major disasters occur. However, manually inspecting pipelines is difficult, slow, and dangerous due to the narrow diameters and hazardous contents. Using mobile robots is an emerging solution, allowing autonomous inspection and mapping of hard-to-reach pipelines without endangering human lives.

The key capabilities needed for such robotic inspection include sensors for obstacle detection, cameras for vision, locomotion mechanisms to move within pipes, on-board power, and wireless communication. SLAM algorithms are essential to build maps of unknown environments while tracking the robot's location. This project explores a ROS-based implementation of SLAM using both simulation and a physical robot testbed. The goal is to demonstrate the feasibility of autonomously mapping an indoor maze as a simplified representation of real-world pipelines.

*A. Prior work*

Several robotic systems have been developed and researched for in-pipe inspection and cleaning, though most cases assume predictable environments and operating conditions [5]. Handling bends and deviations in pipelines adds challenges for mapping and navigation.

Some examples of prior inspection robots include

- A serpentine robot capable of SLAM inside pipes using ultrasound sensors and infrared imaging for short ranges [2].

- Robots with tank-style wheels, as used in this project, for traversing small irregularities [4].

- Robots based on worm-like crawling or wheeled locomotion with cameras and laser scanners for vision [3].

- Robots adhere to pipe walls using magnetic wheels or legs and move by propelling against walls [1].

While progress is being made, practical applications currently focus on large-diameter pipes with low fluid flows due to the difficulties of unknown environments. For reliable inspection, robots need to independently navigate with online mapping and obstacle avoidance capabilities. This project aims to provide a proof-of-concept using ROS-based SLAM on a simple two-wheeled robot.

II. METHODOLOGY

The key challenge for a mobile robot is determining its position and orientation within the environment. GPS can provide sufficient outdoor localization but fails indoors, requiring an alternative solution. If a robot has perfect sensors, it can easily build a map of its surroundings. However, in reality, this is impossible - the robot faces an unknown environment, all sensors have some errors, and uncertainties in mapping accumulate over time [6].

To address this chicken-and-egg problem of mapping requiring localization while localization relies on the map, an iterative mathematical approach known as simultaneous localization and mapping (SLAM) is taken. SLAM incrementally builds a map while simultaneously localizing the robot, propagating and reducing errors over time. This enables autonomous mapping of unknown environments [7].

This project implements a 2D SLAM algorithm on a simulated differential drive robot equipped with sensors. The virtual environment is an indoor maze representing simplified pipeline geometry. The mapper generates a 2D occupancy grid map while outputting an estimate of the robot's changing position. The algorithm and sensors are then implemented on a physical robot in the real maze. A key focus is improving mapping accuracy without wheel encoders through sensor fusion.

III. Demonstration

A. Simulation

Robotic simulations play an important role in robotics research and development by providing a safe and cost-effective testbed for algorithms, hardware designs, and software systems before real-world deployment. In this work, simulations were conducted using the Robot Operating System (ROS) and its integrated 3D simulation platform, Gazebo.

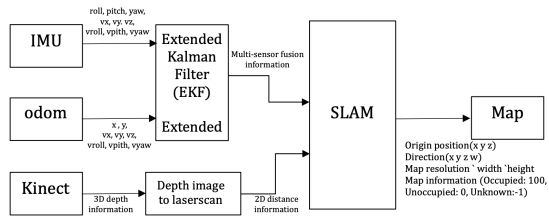

Figure 1: Localization method with IMU, odometry data, and Kinect diagram

A 3D robot model was developed using the SolidWorks CAD software and exported to Universal Robotic Description Format (URDF) for use in Gazebo. This allowed a realistic rendering of the robot's visual appearance and physical properties such as link sizes, masses, and inertial properties.

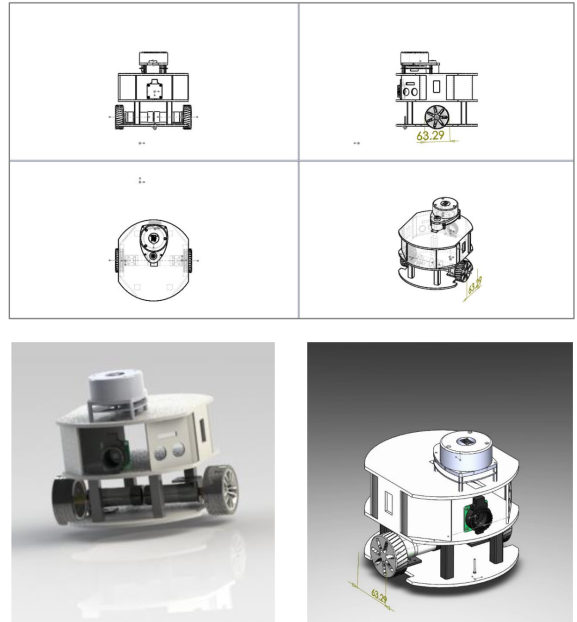

Figure 2: CAD model of the simulated robot

Simulated robotic experiments were performed in a virtual environment modeled within Gazebo. The robotic motions were executed using the ROS navigation stack which integrates control capabilities. This facilitated the evaluation of the overall system performance in a maze.

Odometry data from the simulated tf data and scans from the simulated Lidar sensor were used as input to the Gmapping SLAM algorithm for building maps of the environments. Qualitative analysis through map visualizations provided insights into the mapping quality and how well environmental structures were reconstructed.

In addition, the robot's localization performance was tested using the AMCL module. Trajectories recorded during simulations were played back and the estimated trajectory from AMCL was compared to the ground truth from Gazebo to calculate localization errors over time. This allowed analyzing factors affecting long-term localization drift. Together, the simulation results served to validate and refine the SLAM solution before real-world robot experiments.

### B. Physical Robot Platform

A physical two-wheeled differential drive robot was constructed by integrating off-the-shelf components including Rasberry Pi, a motor controller, and a Kinect camera. The robot chassis was custom-fabricated using laser-cut acrylic parts.

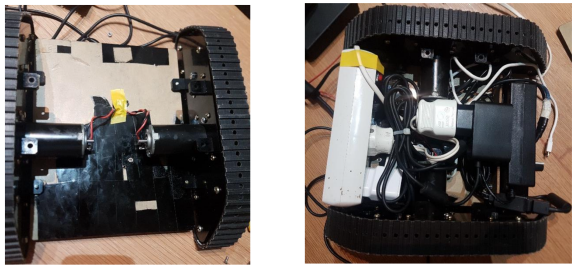

Figure 3: The robot's chassis

An inertial measurement unit (IMU) was added to provide rotation and acceleration data.

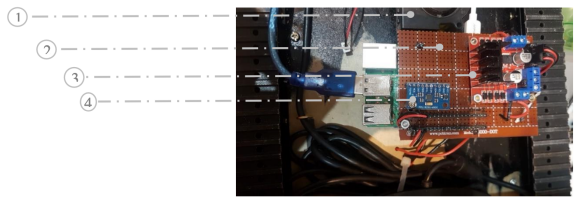

Figure 4: A top-down view of the robot's control system: 1) Fan  2) Control board mounted on the chassis   3) IMU driver (MPU9250 sensor)  4) Motor driver (L298N chip)

## IV. Experiments and Results

The robot was manually driven through a constructed indoor maze environment while capturing sensor data for mapping. The SLAM and navigation algorithms developed in the simulation were implemented on the physical robot using ROS. The sensor data was visualized in RViz and the output map was generated.

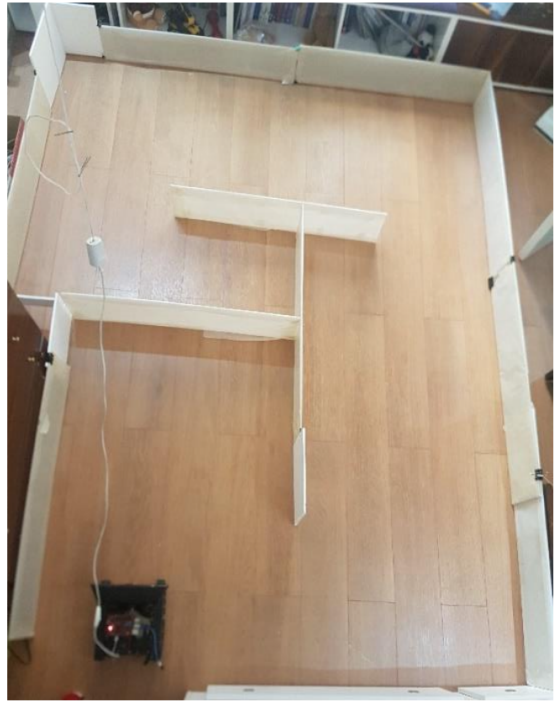

Figure 5: The constructed maze

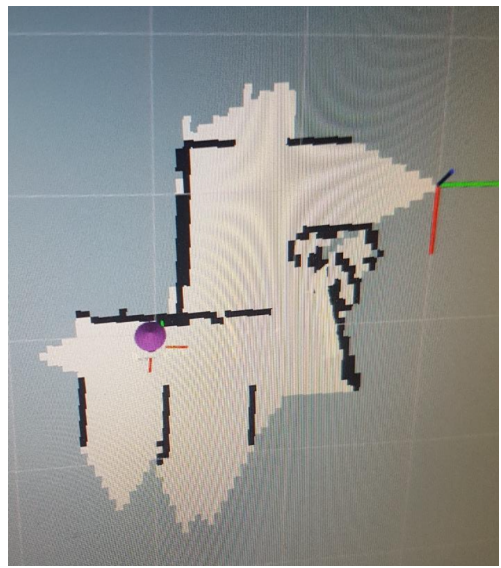

Figure 6: Real-time sensory data visualization on RViz

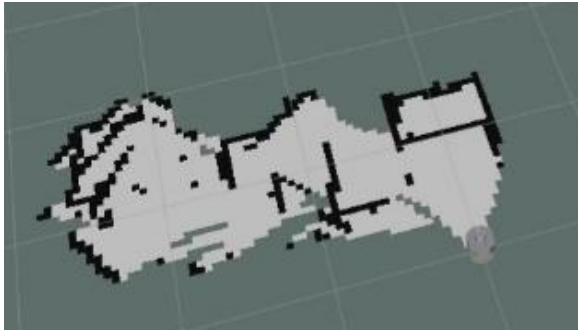

Figure 7: Preliminary Test Results of Map Generation

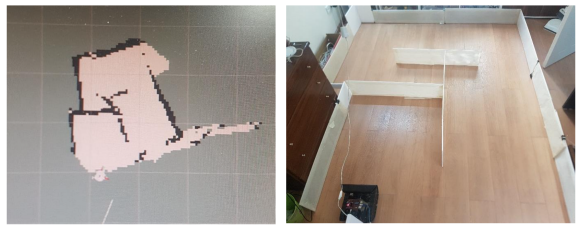

Figure 8: Evaluating the Final Test Map Output Against the Actual Floorplan

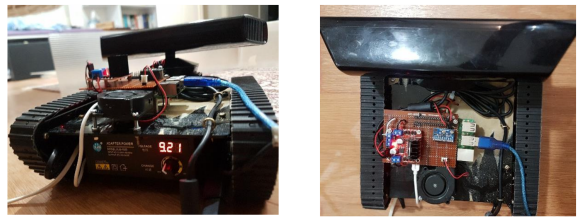

Figure 9: Early design of the robot

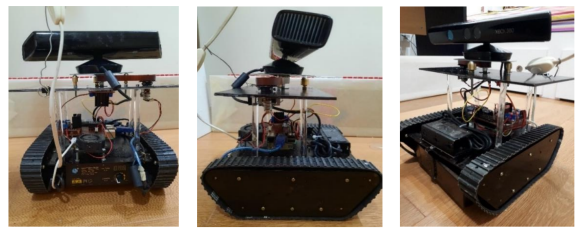

Figure 10: Final design of the robot

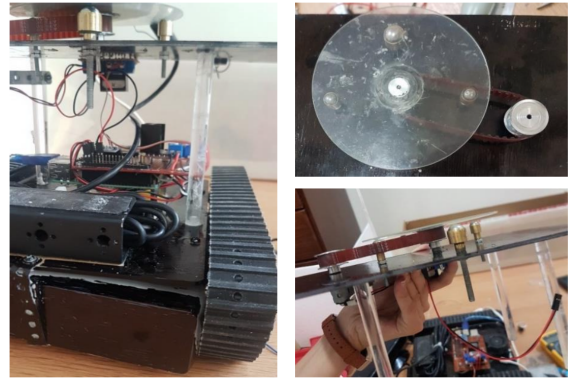

Figure 11: Belt and pulley rotation mechanism for manually driving the Kinect for 360 degrees capturing

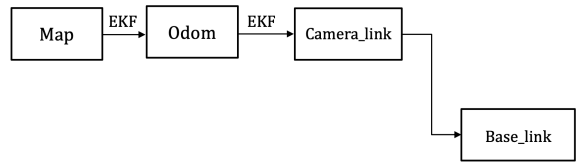

Figure 12: General overview of the frame sequences in fusing sensor data for localization

## V. Contributions

This project provides an overview of applying ROS and SLAM methods to autonomously map an unknown indoor environment representing simplified petrochemical pipeline geometry. The specific contributions include:

- Designing and fabricating a two-wheeled robot chassis to serve as an experimental platform

- Modeling the robot in ROS using URDF for physics simulation

- Implementing SLAM using Gmapping in Gazebo and tuning parameters

- Developing a sensor fusion pipeline to integrate Kinect camera and IMU data for improving odometry [8]

- Testing and benchmarking mapping accuracy with and without fusion on the physical robot

- Providing recommendations for real-world application based on analysis of simulation versus experimental results

The project can inform the future development of practical pipeline inspection robots using ROS-based SLAM approaches.

## VI. Conclusion

This project demonstrated a proof-of-concept application of using ROS and SLAM for autonomously mapping an unknown indoor environment representing simplified petrochemical pipeline conditions. Both simulation and physical robot experiments were conducted to implement 2D occupancy grid mapping using laser scan data and sensor fusion techniques.

The final map produced by manually driving the physical robot through the maze showed reasonable correspondence to the true layout, despite the lack of wheel encoders. This indicates that encoding odometry using visual and inertial data can be a feasible alternative for pipeline mapping robots. However, accuracy was heavily dependent on sensor calibration, algorithm tuning, and environmental conditions.

While the results help validate the approach, substantial further development would be required for real-world deployment in petrochemical plants. Key areas needing improvement include:

- Insulation and waterproofing for hazardous environments

- Accounting for temperature effects, rust, and other factors affecting sensor accuracy

- Handling a wider range of pipe geometries and sizes

- Increased robustness to unpredictable disturbances like fluid flows

- Advanced control algorithms for autonomous navigation and exploration

- Higher resolution sensing for detecting small cracks and leaks

- Adding redundancy for mission-critical uses

The sensor payload and software algorithms would need to be tailored to the specific application requirements. LIDAR scanners can provide more reliable longer-range measurements than the Kinect depth camera used. The SLAM algorithm parameters would also require tuning based on sensor error profiles.

In conclusion, this project provides a valuable starting point for the future development of ROS-based autonomous pipeline inspection robots. The experimental results highlight challenges that would need to be addressed for real-world viability across diverse pipeline conditions. Suggestions are provided on mechanical design, sensor selection, and algorithm improvements to robustly handle industrial environments.